\documentclass[conference]{IEEEtran}

\IEEEoverridecommandlockouts
\usepackage{cite}
\usepackage[numbers,sort&compress]{natbib}

\usepackage[utf8]{inputenc}
\usepackage{amsmath,amssymb}
\usepackage{graphicx}
\usepackage{bm}
\usepackage{enumerate}
\usepackage{comment}
\usepackage{xcolor}
\usepackage{url}
\usepackage{color,soul}
\usepackage{subcaption}
\usepackage{float}
\usepackage{tablefootnote}

\definecolor{light-gray}{gray}{0.8}

\usepackage{amsmath,amssymb,amsfonts}
\usepackage{algorithmic}
\usepackage{graphicx}
\usepackage{textcomp}
\usepackage{xcolor}
\usepackage{subcaption}

\begin{document}

\title{Exploring Diverse Methods in Visual Question Answering \\
}

\author{
\small 

\begin{tabular}[t]{c@{\extracolsep{8em}}c} 

1\textsuperscript{st} Panfeng Li\footnotemark & 2\textsuperscript{nd} Qikai Yang \\
\textit{Department of Electrical and Computer Engineering} & \textit{Department of Computer Science} \\
\textit{University of Michigan} & \textit{University of Illinois Urbana-Champaign} \\
Ann Arbor, USA & Urbana, USA \\
pfli@umich.edu & qikaiy2@illinois.edu \\

\\

3\textsuperscript{rd} Xieming Geng & 4\textsuperscript{th} Wenjing Zhou \\
\textit{Department of Computer Science} & \textit{Department of Statistics} \\
\textit{Chongqing University} & \textit{University of Michigan} \\
Chongqing, China & Ann Arbor, USA \\
xieminggeng@foxmail.com & wenjzh@umich.edu \\

\\
5\textsuperscript{th} Zhicheng Ding & 6\textsuperscript{th} Yi Nian \\
\textit{Fu Foundation School of Engineering and Applied Science} & \textit{Department of Computer Science} \\
\textit{Columbia University} & \textit{University of Chicago} \\
New York, USA & Chicago, USA \\
zhicheng.ding@columbia.edu & nian@uchicago.edu \\

\end{tabular}

}

\maketitle

\begin{abstract}
This study explores innovative methods for improving Visual Question Answering (VQA) using Generative Adversarial Networks (GANs), autoencoders, and attention mechanisms. Leveraging a balanced VQA dataset, we investigate three distinct strategies. Firstly, GAN-based approaches aim to generate answer embeddings conditioned on image and question inputs, showing potential but struggling with more complex tasks. Secondly, autoencoder-based techniques focus on learning optimal embeddings for questions and images, achieving comparable results with GAN due to better ability on complex questions. Lastly, attention mechanisms, incorporating Multimodal Compact Bilinear pooling (MCB), address language priors and attention modeling, albeit with a complexity-performance trade-off. This study underscores the challenges and opportunities in VQA and suggests avenues for future research, including alternative GAN formulations and attentional mechanisms.
\end{abstract}


\section{Introduction}

The Visual Question Answering (VQA) task~\cite{VQA}, represents a pivotal challenge within the field of Artificial Intelligence (AI), striving to bridge the gap between visual perception and natural language understanding. At its core, VQA endeavors to imbue machines with the capacity to comprehend and respond to inquiries pertaining to the content of images, thereby mirroring a fundamental aspect of human cognition. Indeed, the ability to seamlessly integrate visual information with linguistic input is regarded as a hallmark of human intelligence, facilitating the formation of abstract concepts and fostering deeper comprehension of the surrounding environment.

In essence, the VQA task epitomizes the interdisciplinary nature of contemporary AI research~\cite{Li2019JointCA}, drawing upon insights from computer vision~\cite{zhu2021pseudo}, natural language processing~\cite{chen2024taskclip}, and machine learning~\cite{wang2024adapting} to realize the ambitious goal of imbuing machines with human-like perceptual and cognitive abilities. By synthesizing information gleaned from visual stimuli with linguistic cues, VQA systems aspire to emulate the nuanced understanding and reasoning capabilities exhibited by human agents when confronted with multimodal inputs.

In light of the aforementioned considerations, this study embarks on a journey to unravel the intricacies of the VQA conundrum, leveraging insights from one of the most prominent and widely utilized datasets in the field~\cite{balanced}. By delving into the depths of this seminal dataset, we endeavor to shed light on the underlying challenges and opportunities inherent in the VQA paradigm, with the ultimate aim of advancing the frontier of AI research and fostering the development of more intelligent and perceptive machines.

\section{GAN BASED MECHANISM}
Our rationale for applying Generative Adversarial Networks stemmed from several considerations:

\begin{enumerate}
\item VQA is fundamentally a generative task---given an image and question, we must generate an answer. We thus reasoned that GANs, due to their generative nature, may be a good fit for the problem.
\item Conditional GAN \cite{reed} made us reason that, by inputing question and image embeddings as 2-tuples to the generator, we could learn to accurately generate answer embeddings as the outputs. This is because the generator will learn the probability distribution of the answer embeddings conditioned on the question-image embedding 2-tuple. By thinking of the discriminator term as an adaptive loss function, we also reasoned that, with proper training, this may be able to determine a more appropriate loss for the VQA classification problem.
\end{enumerate}

Since the goal of our GAN-based approach was, in part, to simply investigate the application of GANs to classification problems, we focused more on experimenting with and testing a wide variety of possible applications of GANs to this problem than on beating the state-of-the-art results.

\subsection{Architecture}

Similar to most existing approaches to the VQA problem, we first passed the image data through a pre-trained ResNet to generate a feature vector for each image. In parallel, we also pre-processed the questions and pass them through an RNN in order to obtain feature vectors for each question. While we do not train the ResNet, we do train the RNN while training the GAN. After generating the feature vectors, we encode them together into a single vector. We attempted two approaches to combining them. Our first approach, motivated by \cite{saito, huang20}, first passed the image and question embeddings through separate $tanh$ layers, then performed element-wise multiplication of these output vectors, and concatenated them with a vector of the element-wise attention of the outputs of the $tanh$ layers. This vector is finally passed through another $tanh$ nonlinearity. Our second approach was much simpler, utilizing separate $tanh$ layers to encode the image and question feature vectors and then combining these embeddings by element-wise multiplication. 

Once we had generated the combined embeddings of the features, we fed them into our generator network. These feature vectors can be thought of as the conditional input to the generator, allowing it to generate an output conditioned on the image and question. In addition to feeding in the feature embeddings as conditions for the output, we also inputed random noise into the generator. We tested doing this in three different ways---by concatenating the noise to the feature vector (\textbf{N1}), by adding the noise to the feature vector to essentially create a non-zero mean noise input (\textbf{N2}), and by feeding in only the features with no noise (\textbf{N0}). While the latter approach may not be a true GAN, it can nonetheless be trained as a GAN and gave us a baseline to compare against. We tested with two different architectures for the generator. The first utilized three fully connected layers with ReLu nonlinearities followed by a linear fully connected layer. The second simply utilized a single linear layer. Each of the generators output a vector of length 1000 encoding the likelihood of each of the 1000 most common answers.

We coupled the simple, single layer generator with the simpler of our embedding methods---for future reference we denote this network as $\mathbf{G_{simp}}$---and the more complex generator with our more complex embedding method---which we denote as $\mathbf{G_{full}}$.

\begin{figure}[]
    \centering
    \includegraphics[width=.95\columnwidth]{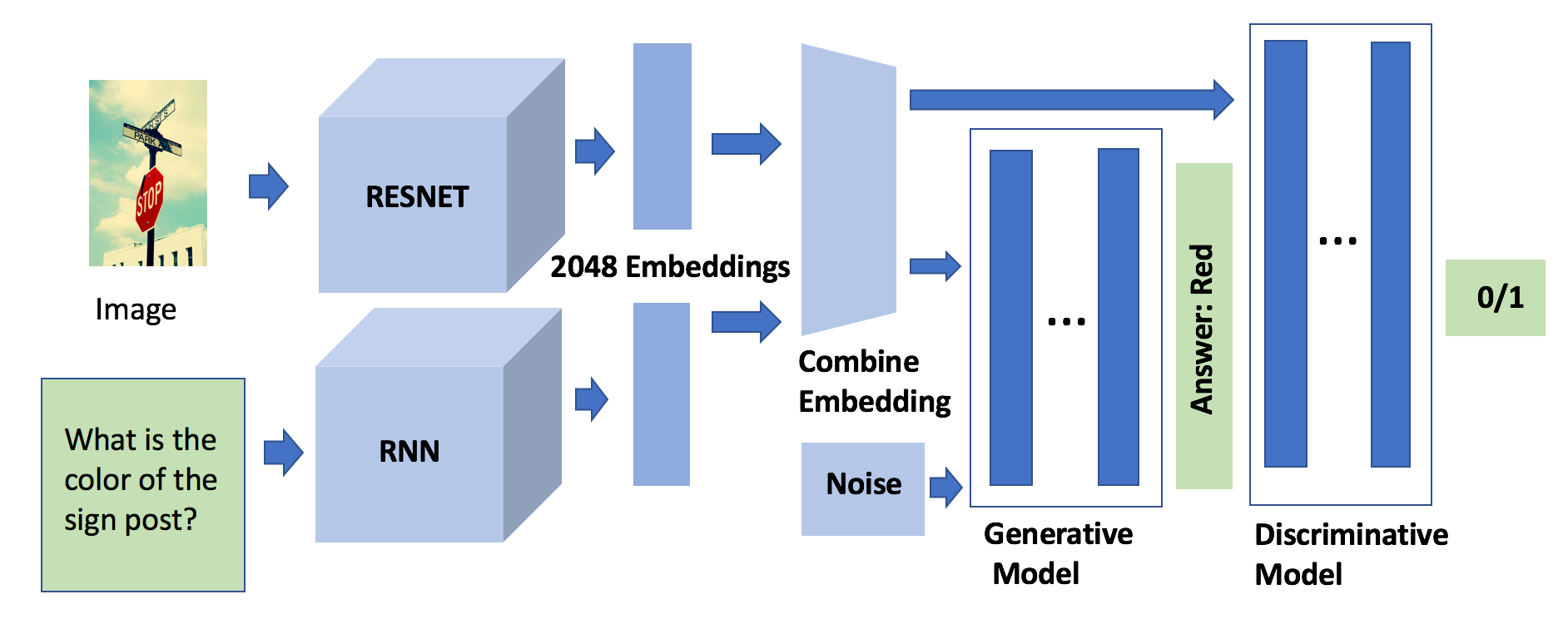}
    \caption{High Level Architecture of GAN Based System}
    \label{gan_architecture}
\end{figure}

We tested with training generators without the discriminator portion and also, in order to produce the full GAN, we fed the generators' outputs into a discriminator network. We attempted several different architectures of the discriminator portion but each was ultimately several fully connected ReLu layers which output a single number into a sigmoid activation to scale it between 0 and 1. In addition to taking the output of the generator as an input, we also fed the features associated with each question and image into the discriminator by concatenating them to the output produced by the generator. We attempted several variations on this, testing with feeding in the raw features produced by the ResNet and RNN and feeding in combined, embedded features. We denote the coupling of the simple generator with a discriminator as $\mathbf{GAN_{simp}}$ and the coupling of the more complex generator with a discriminator as $\mathbf{GAN_{full}}$ A diagram of this architecture is given in Figure \ref{gan_architecture}.

\subsection{Training}

Our basic training algorithm followed the GAN-CLS with step size $\alpha$, outlined by Reed \textit{et. al.}:
\begin{enumerate}
\item \textbf{Input}: minibatch answer embeddings $x$, matching image-question embedding 2-tuple $\psi(t)$, number of training batch step $S$
\item \textbf{for} $n=1$ \textbf{to} $S$ \textbf{do}
\item $h \leftarrow \psi(t)$ (Encode image-question tuple that matches with the answer)
\item $\hat{h} \leftarrow \psi(\hat{t})$ (Encode image-question tuple that doesn't match with the answer)
\item $z \sim \mathcal{N} (0,1)^Z$ (Draw random noise)
\item $\hat{x} \leftarrow G(z,h)$ (Forward through Generator)
\item $s_r \leftarrow D(x,h)$ (correct answer, correct tuple)
\item $s_w \leftarrow D(x,\hat{h})$ (correct answer, wrong tuple)
\item $s_f \leftarrow D(\hat{x},h)$ (output of Generator, correct tuple)
\item $\mathcal{L}_D \leftarrow log(s_r) + \frac{1}{2}(log(1-s_w)+log(1-s_f))$
\item $D \leftarrow D- \alpha \frac{\delta \mathcal{L}_D} {\delta D}$ (Update Discriminator)
\item $\mathcal{L}_G \leftarrow log(s_f)$
\item $G \leftarrow G- \alpha \frac{\delta \mathcal{L_G}} {\delta G}$ (Update Generator)
\item \textbf{end for}
\end{enumerate}

We attempted several variations of this. We experimented with pre-training both the Discriminator and the Generator. When pre-training the Generator we simply trained it as a softmax classifier with noise added to the inputs. This allows us to initialize the weights to values that will produce relatively good results at the start of training the full GAN~\cite{deng2023plgslam}.

We found that pre-training the Discriminator to optimality could be detrimental to the actual training process of the GAN and could worsen the updates of the Generator over time \cite{pretrain}. \cite{pretrain} shows that if the probability densities are either disjoint or lie on a low-dimensional manifold then the Discriminator can distinguish between them perfectly. This happens to be the case for the loss functions proposed by \cite{goodfellow}. So instead of pre-training our Discriminator to optimality, we follow the suggestion in \cite{yao2023ndc} and add noise to its inputs.

We tested adding normalization to the outputs of the layers in our generator and discriminator modules. In general, however, we found that this yielded worse results than when using unnormalized layers. We utilized a small amount of dropout in training our generator and discriminator modules.

We attempted initializing the weights under several different distributions, noticing altered results when we did so. We first initialized all weights to follow a Gaussian distribution with large values clipped (we denote this initialization method $\mathbf{I1}$) and also tested with initializing weights to follow a uniform distribution (which we denote $\mathbf{I2}$).


\section{AutoEncoder Based Mechanism}
We modified the initial GAN technique to give us an Autoencoder based technique, wherein the concatenated features are passed through an autoencoder to generate low dimensional embeddings. Most existing approaches (such as MCB \cite{mcb}) utilize a fixed method to embed the question and image features together. By employing an autoencoder, we hoped to learn how to best embed the question and image features into a low-dimensional space. We use this encoding, after passing it through several fully connected layers, to generate the answer.

\begin{figure}[H]
\centering
\includegraphics[width=.95\columnwidth]{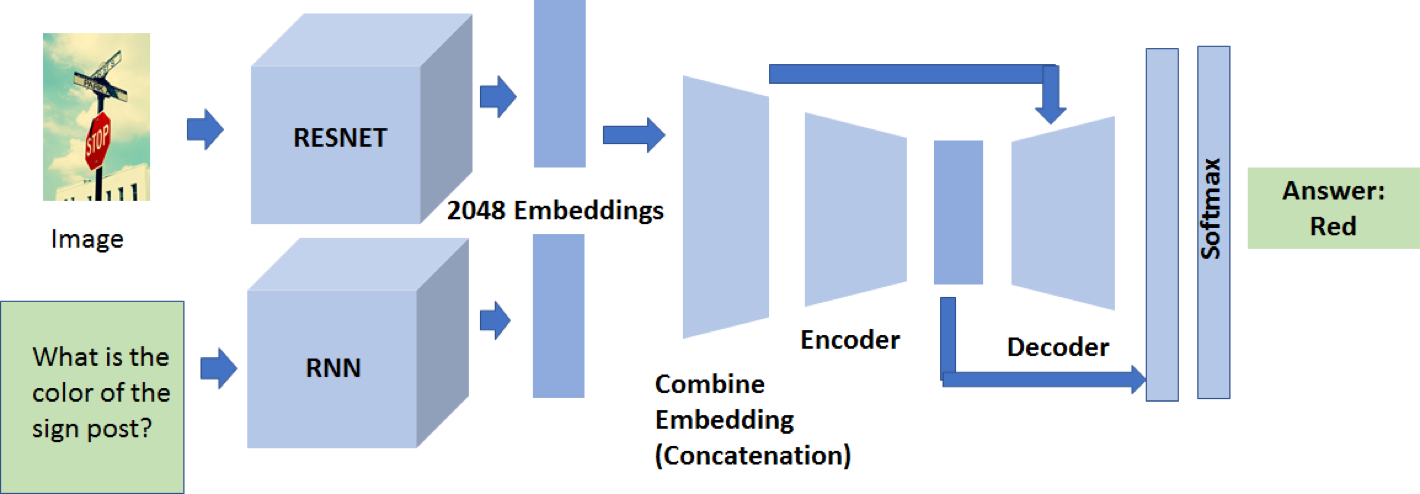}
\caption{High Level Architecture of AutoEncoder Based System}
\label{auto_architecture}
\end{figure}

\section{Attention Based Mechanism}
To answer a question according to an image, it is critical to model both “where to look” and model “what words to listen to”, namely visual attention and question attention \cite{co-attention}. 

However, \cite{balanced} shows that the language priors make the VQA dataset \cite{VQA} unbalanced, whereas simply answering “tennis” and “2” will achieve 41\% and 39\% accuracy for the two types of questions--- “What sport is” and “How many”. These language priors bring to light the question of whether machines truly understand the questions and images or if they tend to give an answer which has higher frequency in the dataset. 

Inspired by the strength of Multimodal Compact Bilinear Pooling (MCB) at efficiently and expressively combining multimodal features \cite{mcb}, we use the MCB operation to replace the simple addition operation used in co-attention mechanism \cite{co-attention} when combining the features learned from the images and questions together, which may help to learn more information from visual part~\cite{ding18, ding20}. 


\begin{figure}[H]
\centering
\includegraphics[width=.95\columnwidth]{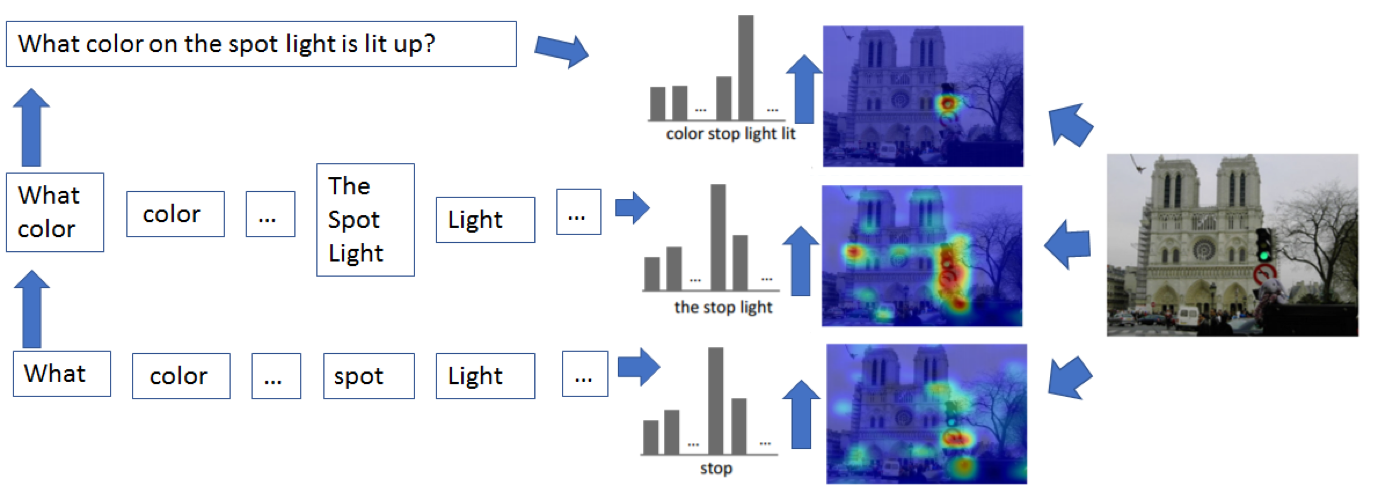}
\caption{High Level Architecture of Attention Based System}
\label{att_architecture}
\end{figure}


\section{Results}


Table \ref{allresults} gives the numerical results obtained by every method we tested. The metric used is the one presented in \cite{VQA} where an answer is considered correct and given a score of one if at least three of the ten human given responses match that answer. As this table illustrates, the baseline methods in general do worse than all novel approaches we attempted.

\begin{table}
\begin{center}
\begin{tabular}{ |c|c|c|c|c| } 
 \hline

 \textbf{Method} & \textbf{All} & \textbf{Yes/No} & \textbf{Number} & \textbf{Other} \\ 
 \hline
 \multicolumn{5}{|c|}{\textbf{Baseline Methods}} \\
 \hline
  $\mathbf{G_{simp} - N0 - I1}$ & 11.58 & 23.86 & 5.32 & 1.48 \\ 
 \hline
 $\mathbf{G_{full} - N0 - I1}$ & 18.65 & 40.56 & 0.25 & 7.84 \\ 
 \hline
 $\mathbf{G_{full} - N1 - I1}$ & 23.41 & 55.46 & 3.49 & 0.61 \\ 
 \hline
 $\mathbf{G_{full} - N2 - I1}$ & 14.76 & 35.77 & 1.25 & 0.27 \\ 
 \hline
 $\mathbf{G_{full} - N2 - I2}$ & 23.06 & 54.65 & 3.50 &  0.51 \\ 
 \hline

  \multicolumn{5}{|c|}{\textbf{Novel Methods}} \\
  \hline

 $\mathbf{GAN_{simp} - N0 - I1}$ & 25.27 & 62.49 & 0.32 & 0.59 \\
 \hline
 $\mathbf{GAN_{simp} - N0 - I2}$ & 25.57 & 56.67 & 9.05 & 0.61 \\
 \hline
 $\mathbf{GAN_{full} - N0 - I1}$ & 27.51 & 51.90 & 22.41 & 0.08 \\
 \hline
 $\mathbf{GAN_{full} - N1 - I1}$ & 28.81 & 57.28 & 19.13 & 0.54 \\
 \hline
 $\mathbf{GAN_{full} - N2 - I1}$ & 34.57 & 65.38 & 27.36 & 0.71 \\
 \hline
 $\mathbf{Autoencoder}$ & 37.65 & 64.01 & 24.37 & 15.77  \\
 \hline
 $\mathbf{Attention}$ & 44.32 & 66.64 & \textbf{32.15} & 26.72 \\
 \hline
 $\mathbf{Attention + MCB}$ & \textbf{47.58}& \textbf{67.60} & 31.47 & \textbf{36.98} \\
 \hline
\end{tabular}

\caption{Results on VQA 1.9 Validation Dataset. Legend: $\mathbf{G_{simp}}$ - simple, single-layer generator trained as classifier; $\mathbf{G_{full}}$ - full, multi-layer generator trained as classifier; $\mathbf{GAN_{simp}}$ - simple, single-layer generator trained with discriminator; $\mathbf{GAN_{full}}$ - full, multi-layer generator trained with discriminator; $\mathbf{N1}$ - noise concatenated to generator conditioning input; $\mathbf{N2}$ - noise added to generator condition input; $\mathbf{N0}$ - no noise inputed to generator; $\mathbf{I1}$ - weights initialized via Gaussian distribution; $\mathbf{I2}$ - weights initialized via uniform distribution}

\label{allresults}
\end{center}
\end{table}

\begin{figure*}
\centering
\begin{subfigure}{0.22\textwidth}
\includegraphics[width=1\columnwidth, scale=0.6]{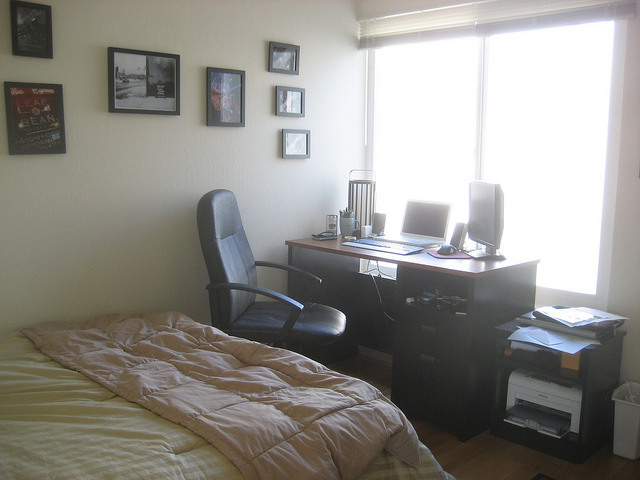}
\caption{Question: How many chairs are in the photo? Baseline Answer: 3. GAN Answer: 1. Attention Answer: 1}
\end{subfigure}
~
\begin{subfigure}{0.22\textwidth}
\includegraphics[width=1\columnwidth, scale=0.6]{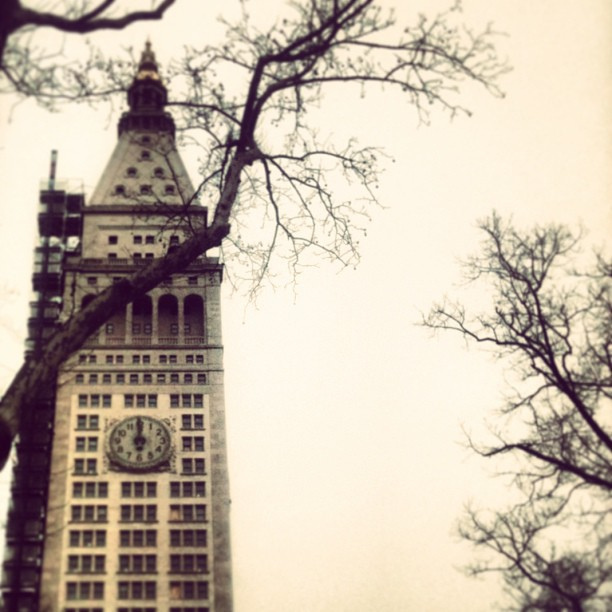}
\caption{Question: Is it an overcast day? Baseline Answer: yes. GAN Answer: yes. Attention Answer: yes}
\end{subfigure}
~
\begin{subfigure}{0.22\textwidth}
\centering
\includegraphics[width=0.6\columnwidth, scale=0.6]{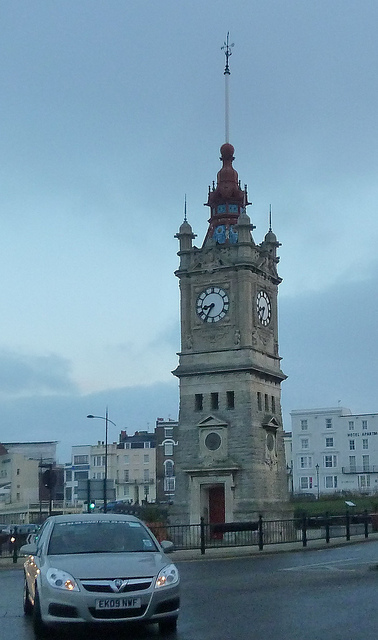}
\caption{Question: What year is the car? Baseline Answer: scarf. GAN Answer: 2010. Attention Answer: 2010}
\end{subfigure}
~
\begin{subfigure}{0.22\textwidth}
\includegraphics[width=0.8\columnwidth, scale=0.6]{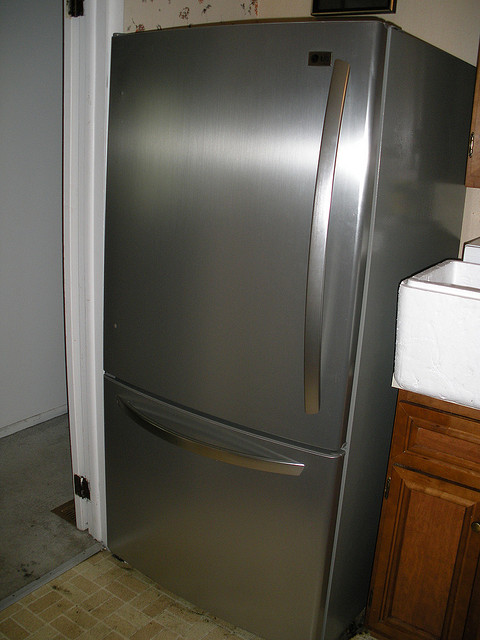}
\caption{Question: What color is the fridge? Baseline Answer: dinner. GAN Answer: gray. Attention Answer: silver }
\end{subfigure}
\caption{Qualitative Results of Visual Question Answering}
\label{qualitative}
\end{figure*}

Figure \ref{qualitative} illustrates the qualitative performance of one of our baseline approaches, one of the GAN-based approaches, and the attention-based model on several images. While the baseline model is able to correctly answer the simpler questions, it fails miserably at more complex questions such as (c) and (d). For (d) it seems to capture some of the meaning of the question (relating a fridge to a meal) yet still incorrectly answers the prompt. As for the GAN-based approach and the attention-based approach, both are able to correctly capture the meaning of the questions while the attention-based approach is slightly better than the GAN-based approach - in (d) our attention-based approach gets the correct answer while our GAN-based approach returns an almost correct answer.

Table \ref{results_pretraining} illustrates the effect various pretraining combinations have on the results produced by the GAN. There results are given on the GAN utilizing the full, multi-layer generator model $\mathbf{GAN_{full}}$ with noise added to the conditioning inputs $\mathbf{N2}$ and weights initialized according to a Gaussian distribution $\mathbf{I1}$. From this it is clear that the optimal results are obtained when the generator is pretrained but the discriminator is not. When we do not pretrain the generator at all, we found that, given the amount of time for which we trained, the generator is unable to learn the distribution of the data at all. In contrast, discriminator pretraining alone doesn't enhance performance.

\begin{table}
\begin{center}
\begin{tabular}{ |c|c|c|c|c| } 
 \hline
 \textbf{Method} & \textbf{All} & \textbf{Yes/No} & \textbf{Number} & \textbf{Other} \\ 
 \hline
 Neither Pretrained & 0.10 & 0.08 & 0.19 & 0.10 \\ 
 \hline
 G Pretrained & \textbf{21.25} & \textbf{54.65} & \textbf{3.50} & \textbf{0.51} \\ 
 \hline
 D Pretrained & 0.05 & 0.00 & 0.09 & 0.08 \\ 
 \hline
 Both Pretrained & 13.76 & 35.47 & 2.16 & 0.30 \\ 
 \hline
\end{tabular}
\end{center}
\caption{Varying Pretraining on GAN with $\mathbf{N2}$ (noise added) on VQA 1.9 Validation Dataset}
\label{results_pretraining}
\end{table}

\section{Conclusion}
This research has embarked on a comprehensive exploration of advanced techniques in VQA, using GANs, attention mechanisms, and autoencoders to enhance the performance of VQA systems. Based on the results from the balanced VQA dataset, we can draw several important conclusions about the effectiveness of these methods in improving VQA systems.

Firstly, our experiments with GANs, especially those with full, multi-layer generator models ($GAN_{full}$), showed a marked improvement over baseline methods when pretraining is applied selectively—specifically, pretraining the generator but not the discriminator. This approach allowed the GANs to better capture and reproduce the data distribution needed for accurate VQA, although it still presents challenges in task specificity and learning the correct answers to more complex questions, aligned with previous studies~\cite{song23c_interspeech}.

Secondly, our investigation into autoencoder-based techniques revealed that while they are effective at learning optimal embeddings for the question and image data, they achieved slightly better results compared to the GAN-based approach. This is largely due to the much better results on more complex questions. On the contrast, our optimized $GAN_{full}$ method yields better on Yes/No and Number questions.

Lastly, the attention mechanisms, particularly those employing MCB, have demonstrated a substantial benefit in addressing the inherent language priors and improving the modeling of attention over both the textual and visual inputs. This method outperformed both the GAN-based and autoencoder-based approaches in complex question answering scenarios, as seen in the qualitative performance comparisons. However, this comes with the cost of increased computational complexity, which poses a trade-off that future research will need to address.

Overall, the research indicates that while the innovative approaches explored—GANs, autoencoders, and attention mechanisms—show promise, they each come with unique challenges that need to be overcome. Future research should focus on refining these methods, perhaps through alternative GAN formulations and enhanced stability techniques for more complex tasks. Additionally, further exploration into hybrid models combining these techniques could yield improvements in both performance and efficiency for VQA systems. This study thus not only advances our understanding of complex VQA systems but also outlines clear paths for future developments in the field.

\renewcommand{\bibfont}{\footnotesize}

\footnotesize{
\bibliographystyle{IEEEtran}
\bibliography{main}
}

\end{document}